\documentclass[sigconf]{acmart}
\AtBeginDocument{%
  }

\setcopyright{acmlicensed}
\copyrightyear{2025}
\acmYear{2025}
\acmDOI{X}
\acmConference[]{SC' 25}{St. Louis, MO, USA, Nov. 16-21, 2025}

\acmISBN{978-1-4503-XXXX-X/2018/06}

\usepackage{pifont}
\usepackage{multirow}
\usepackage{graphicx}

\begin{document}

\title{MoE-Compression: How the Compression Error of Experts Affects the Inference Accuracy of MoE Model?}



\author{Songkai Ma}
\affiliation{%
  \institution{Department of Computing}
  \institution{Hong Kong Polytechnic University}
  \country{Hong Kong}
}
\email{songkai.ma@connect.polyu.hk}

\author{Zhaorui Zhang}
\authornote{Zhaorui Zhang is the corresponding author. This paper has been accepted by the workshop of SC'25.}
\affiliation{%
  \institution{Department of Computing}
  \institution{Hong Kong Polytechnic University}
  \country{Hong Kong}}
\email{zhaorui.zhang@polyu.edu.hk}

\author{Sheng Di}
\affiliation{%
  \institution{Mathematics and Computer Science Division}
  \country{Argonne National Laboratory, USA}
}
\email{sdi1@anl.gov}

\author{Benben Liu}
\affiliation{%
  \institution{LSCM R\&D Center}
  \institution{The University of Hong Kong}
  \country{Hong Kong}}
\email{benbenliu@hku.hk}

\author{Xiaodong Yu}
\affiliation{%
 \institution{Department of Computer Science}
 \institution{Stevens Institute of Technology}
 \country{USA}}
\email{xyu38@stevens.edu}


\author{Xiaoyi Lu}
\affiliation{%
  \institution{Department of Computer Science and Engineering}
  \country{University of California, Merced, USA}}
\email{xiaoyi.lu@ucmerced.edu}


\author{Dan Wang}
\affiliation{%
  \institution{Department of Computing}
  \institution{Hong Kong Polytechnic University}
  \country{Hong Kong}}
\email{dan.wang@polyu.edu.hk}

\renewcommand{\shortauthors}{Zhaorui Zhang et al.}

\begin{abstract}
With the widespread application of Mixture of Experts (MoE) reasoning models in the field of LLM learning, efficiently serving MoE models under limited GPU memory constraints has emerged as a significant challenge. Offloading the non-activated experts to main memory has been identified as an efficient approach to address such a problem, while it brings the challenges of transferring the expert between the GPU memory and main memory. We need to explore an efficient approach to compress the expert and analyze how the compression error affects the inference performance. 

To bridge this gap, we propose employing error-bounded lossy compression algorithms (such as SZ3 and CuSZp) to compress non-activated experts, thereby reducing data transfer overhead during MoE inference. We conduct extensive experiments across various benchmarks and present a comprehensive analysis of how compression-induced errors in different experts affect overall inference accuracy. The results indicate that experts in the shallow layers, which are primarily responsible for the attention mechanism and the transformation of input tokens into vector representations, exhibit minimal degradation in inference accuracy when subjected to bounded errors. In contrast, errors in the middle-layer experts, which are central to model reasoning, significantly impair inference accuracy. Interestingly, introducing bounded errors in the deep-layer experts, which are mainly responsible for instruction following and output integration, can sometimes lead to improvements in inference accuracy.

\end{abstract}



\keywords{Model Compression, Mixture of Experts, Inference, Error Sensitivity}

\maketitle

\vspace{-8pt}
\section{Introduction}

In recent years, Mixture of Experts (MoE) foundation models have enabled large language models (LLMs) to transition from dense architectures to sparsely activated expert frameworks, as exemplified by models such as DeepSeek \cite{guo2025deepseek}, GPT-4 \cite{achiam2023gpt}, Phi-4 \cite{abdin2024phi}, and Mixtral \cite{jiang2024mixtral}. These sparse foundation models selectively activate only a subset of experts for each token, thereby substantially reducing computational overhead and inference costs while maintaining strong generative performance. However, the presence of numerous idle, non-activated experts during MoE inference poses a significant challenge to efficient GPU memory utilization, complicating the deployment of MoE models on GPUs with limited memory resources. For instance, serving the Mixtral-8x7B model requires approximately 94 GB of VRAM when using FP16 precision. In this scenario, only about $30\%$  of the parameters—approximately 27.5 GB—are actively used during the decoding process, while the remaining 66.6 GB of memory is occupied by non-activated experts, resulting in considerable inefficiency \cite{shin2025sparseinfer}. Consequently, it is crucial to address the challenge of improving memory utilization efficiency, as this has a substantial impact on system performance in serving tasks for MoE foundation models.

Offloading techniques \cite{sarkar2023edge, eliseev2023fast, hwang2024pre, song2024promoe, xue2024moe, tang2024hobbit, wang2025storellm}, which transfer expert parameters from main memory to GPU memory on demand for each input, have been recognized as effective solutions to address GPU memory limitations. However, while offloading alleviates memory constraints, it shifts the bottleneck of MoE inference from being memory-bound to I/O-bound. This is primarily due to the need to transfer large volumes of parameters over the relatively low-bandwidth PCIe bus, resulting in substantial data transfer delays. For example, the DRAM-to-VRAM bandwidth provided by PCIe 4.0 (32 GB/s) is orders of magnitude lower than the bandwidth between GPU memory and on-chip computation units (300 GB/s). As a result, existing MoE inference systems employing expert offloading strategies for GPU memory-constrained scenarios \cite{kong2023swapmoe, eliseev2023fast, hwang2024pre, tang2024hobbit} continue to experience sub-optimal performance, with loading delays that are perceptible to users and cannot be effectively masked by concurrent computation tasks. Although low-bit quantization strategies can reduce the size of transmitted parameters and mitigate the latency associated with loading activated experts \cite{eliseev2023fast, sarkar2023edge}, they often lead to significant degradation in generative performance. Therefore, it is essential to investigate efficient compression algorithms that can achieve high compression ratios while maintaining minimal compression error, thereby preserving the generative performance of MoE models. To achieve this goal, there are three critical steps. Our work mainly focuses on the \textbf{first and second steps}.

$\bullet$ \textbf{\textit{Firstly}, we need to investigate an efficient compression algorithm for expert parameters compression during inference that can achieve a high compression ratio while maintaining minimal compression error. }
    
$\bullet$ \textbf{\textit{Secondly}, a comprehensive analysis of compression error sensitivity across different experts is necessary to understand its impact on the generative performance of MoE models. It is important to evaluate how compression errors associated with individual experts influence the overall generation quality. Such analysis will provide valuable insights for designing efficient compression algorithms that minimize offloading overhead while preserving model performance. }
    
$\bullet$ \textbf{\textit{Thirdly}, further enhancements to system performance are required when integrating compression algorithms into the MoE inference framework. In particular, it is important to investigate the design of pipeline algorithms that can overlap compression and decompression operations with offloading tasks, thereby minimizing the associated latency and improving overall inference efficiency.}
    

Currently, four primary expert compression strategies are employed in the context of MoE models: expert distillation \cite{liu2024survey, he2024expertflow}, expert pruning \cite{liu2024survey, lu2024not}, expert decomposition \cite{huang2025milo}, and expert quantization \cite{duanmu2025mxmoe,hu2025moequant, xie2025automated}. Expert distillation involves transferring or compressing the knowledge from a large MoE model with multiple experts into a smaller, more deployable model—typically a single model or a reduced set of experts—thereby preserving performance while reducing computational resource requirements. Expert pruning seeks to optimize resource utilization and reduce redundancy by identifying and removing experts that contribute minimally or perform suboptimally during training. The central principle of expert pruning is to retain experts that significantly enhance model performance and eliminate those whose contributions are limited or whose computational overhead is disproportionately high. Expert decomposition commonly utilizes low-rank decomposition techniques to reduce parameter count by factorizing the weight matrices of MoE models into products of lower-rank matrices. Expert quantization aims to reduce computational and storage costs by decreasing the bit-width of model parameters. By converting floating-point parameters to low-precision integers, quantization can substantially reduce model size and accelerate inference, while striving to maintain model performance. In practice, expert quantization is often combined with other optimization techniques—such as distillation, pruning, and decomposition—to further enhance the efficiency and resource utilization of MoE models during deployment. However, low-bit quantization frequently leads to significant degradation in generative performance for MoE inference tasks, due to the uncontrollable and unpredictable errors introduced during the quantization of expert parameters.

Error-bounded lossy compression approaches \cite{di2025survey, liang2022sz3, huang2023cuszp, huang2024cuszp2, huang2023c, huang2024optimized, huang2025zccl, zhang2025fedcspc, liu2025ocelot, zhang2025fedefsz, zhang2021sapus, zhang2022momentum, zhang2022mipd, xu2024fedfa, zhang2025cllora, liu2025hlora} offer a promising solution for compressing expert parameters with high compression ratios while maintaining minimal compression error, thereby preserving the generative performance of MoE models. These techniques have proven effective in substantially reducing data storage and transfer burdens, all while maintaining high fidelity in the reconstructed data. Numerous error-bounded lossy compressors have been developed to support a wide range of parallel and distributed computing scenarios, each employing distinct compression models and principles that confer specific advantages and limitations. The primary motivation for adopting error-bounded lossy compression for expert parameter reduction is its ability to guarantee a bounded error range, in contrast to traditional quantization methods, which often introduce uncontrollable and unpredictable errors. Furthermore, various error-bounded lossy compression algorithms have been optimized for different hardware platforms and accelerators, including CPU-based algorithms such as SZ3 \cite{liang2022sz3} and GPU-based algorithms such as CuSZp \cite{huang2023cuszp, huang2024cuszp2}.


To the best of our knowledge, this work represents the first attempt to leverage error-bounded lossy compression techniques for compressing expert parameters, with the aim of reducing PCIe offloading overhead and enhancing GPU memory efficiency in MoE inference tasks. As outlined above, achieving these objectives involves three key steps. This study primarily focuses on the first two: \ding{172} investigating efficient compression algorithms that can achieve high compression ratios while maintaining minimal compression error, thereby preserving the generative performance of MoE models during inference; and \ding{173} conducting a comprehensive sensitivity analysis of compression errors across different experts to determine their impact on the overall generative performance of MoE models in inference scenarios. We introduce varying error bounds to add the errors in the experts during inference and evaluate the resulting generative performance using several of the most widely adopted MoE-based foundation models. The main \textbf{contributions and key findings} of this work are summarized as follows:


\ding{182} To the best of our knowledge, this work is the first to propose the use of error-bounded lossy compression algorithms, such as SZ3 and CuSZp, for compressing non-activated experts in order to reduce the data transfer overhead between GPU memory and main memory during inference. In comparison to quantization-based approaches, error-bounded lossy compression algorithms offer higher compression ratios while maintaining minimal compression error, thereby better preserving model performance.
    
\ding{183} We conduct comprehensive experiments using popular MoE models and benchmark datasets to provide an in-depth analysis of the impact of compression error on inference accuracy.
    
\ding{184}  The experimental results indicate that experts in the shallow layers primarily handle attention mechanisms and the transformation of input tokens into vector representations; introducing errors at this stage has only a minimal impact on inference accuracy. In contrast, experts in the middle layers are chiefly responsible for core model reasoning, and the presence of errors in these layers significantly degrades inference accuracy. Interestingly, experts in the deep layers are mainly involved in instruction following and output integration, where the introduction of bounded errors can, in some cases, lead to improvements in inference accuracy.
    

\section{Background and Motivations}

\subsection{MoE Inference Process}
Similar to LLMs, a typical layer in MoE models comprises a self-attention layer followed by a sparse MoE layer. The input tokens to each layer are initially processed by the self-attention mechanism, which can generally be divided into three stages: \ding{172} pre-attention stage, including $QKV$ (query, key, value) projection; \ding{173} self-attention computation stage, involving the calculation of $QK^{T}$; and \ding{174} post-attention stage, which includes output projection. Following the self-attention layer, tokens are passed to the sparse MoE layer, where a router assigns each token to a subset of experts, typically employing a top$-k$ selection strategy \cite{xu2025moe}. Each token is processed by $k$ selected experts, and the final output is obtained by computing a weighted average of the outputs from these experts. Certain model architectures, such as DeepSeek-V2 \cite{liu2024deepseek} and Qwen2MoE \cite{team2024qwen2}, incorporate a shared expert through which all tokens are processed. The resulting token representations are then forwarded to subsequent layers in the model. While layer normalization and residual connections are commonly present in MoE models, they are not central to the discussion here and are therefore omitted.


MoE batched inference closely mirrors the generative inference procedure of LLMs, operating in two distinct phases: \ding{172} \textbf{Prefill:} during which a batch of prompts is processed to generate the key-value (KV) cache at each attention layer; and \ding{173} \textbf{Decoding:} where new tokens are generated in an auto-regressive manner. In the decoding phase, the output tokens from the previous forward pass serve as the input for generating the next token. With each forward pass, the KV-cache corresponding to the new input token is generated and appended to the existing KV-cache, thereby constructing the complete context up to that point. Notably, the computational intensity during the decoding phase is typically orders of magnitude lower than in the prefill phase, as only a single token per sequence is processed by the model at each step.

\subsection{Expert Offloading in MoE Inference Process}

The expert offloading strategy involves the management of two levels of memory: main memory, which stores excess model parameters (weights) and key-value (KV) states, and GPU memory, which is utilized for computation and rapid data access. When model parameters are needed for GPU computation, they can either be prefetched in advance—overlapping with other computations—or fetched on demand. To facilitate efficient data access, a resident store can be implemented in GPU memory to persistently hold model parameters and the KV-cache, while a staging buffer is employed to prefetch dynamic data. If the GPU attempts to access data, including model parameters or KV states, that are not yet present in its resident store, computation must stall until the required data are transferred from main memory. In this context, the bandwidth of the PCIe interface between the host and GPU memory often constitutes a critical bottleneck. Recent research suggests that leveraging CPU computational resources to process a portion of the data locally represents a promising direction, with the potential to increase overall system throughput \cite{cao2025moe}.


\subsection{Motivations from Quantization Approaches}

We summarize various quantization methods in Tab. \ref{quantization_compare}. Our findings indicate that quantization primarily reduces memory usage, with most methods achieving approximately a $4\times$ decrease. While some quantization techniques also accelerate inference, others do not; for instance, QMOE incurs an additional $5\%$  computational overhead. Moreover, our analysis of quantization as a form of lossy compression reveals that lower bit widths generally result in greater performance degradation. Consequently, it is essential to strike a balance among memory efficiency, inference accuracy, and computational speed when applying quantization. However, most existing quantization approaches are unable to simultaneously satisfy these requirements. In this work, we explore a novel direction by incorporating error-bounded lossy compression into the inference process to compress expert parameters, aiming to achieve an optimal balance among memory usage, accuracy, and acceleration.


\begin{table}[!h]
\centering
\caption{The Comparison for Quantization Approaches}
\vspace{-8pt}
\begin{tabular}{c|c c c c}
\toprule[1pt]
\textbf{Method} & \textbf{Mem\_Save} & \textbf{Acc\_Drop} & \textbf{Speedup} & \textbf{Bits} \\
\midrule[0.8pt]
\textbf{MC-MoE} & 4.27$\times$ & 3.8\% & 1.80$\times$ & 1, 2, 3 \\
\textbf{MoE-CSP} & 4.00$\times$ & - & 26.00$\times$ & 4, 8 \\
\textbf{MoQE} & 4.90$\times$ & 0.97\% & $-5\times$ & 2, 3, 4 \\
\textbf{QMoE} & 20$\times$ & 6.7\% & 0.95x & 1, 2 \\
\textbf{CMoE} & 150$\times$ & 23.81\% & - & 1, 2, 4 \\
\textbf{MoE-MPTQS} & - & 4.98\% & $\uparrow$ 20.63$\times$ & 4, 8 \\
\textbf{HOBBIT} & - & 1\% & 1.35$\times$ & 2, 4 \\
\textbf{EdgeMoE} & $\uparrow$1.18$\times$ & 5\% & $\uparrow$ 2.78$\times$ & 2, 4, 8 \\
\bottomrule[1pt]
\end{tabular}
\vspace{-8pt}
\label{quantization_compare}
\end{table}

\section{Methodology for Error Sensitivity Analyze}

\subsection{Benchmark and Experimental Design}

\textbf{MoE Model and Datasets.} We first deployed the Moonlight \cite{liu2025muon} model, whose MoE architecture comprises 26 expert layers, each containing 64 expert submodules. During inference, each layer activates 6 experts, selected via a top$-k$ routing mechanism, with each expert assigned distinct parameter values. For the reasoning tasks, we utilized the GSM8K dataset \cite{cobbe2021training} as input.


\textbf{Errors. } To simulate the compression errors of most current state-of-the-art compressors, such as SZ3 \cite{liang2022sz3}, CuSZp \cite{huang2023cuszp, huang2024cuszp2}, etc., we randomly generated $n$ errors which follows the normal distributions $N \sim (0, \hat{e})$ and add these errors to the expert parameters during MoE inference, where $n$ is the number of parameters that we try to analyze, $\hat{e}$ indicate the error bound of the compression algorithms.

\begin{figure}[!h]
    \centering
    \includegraphics[width=0.98\linewidth, height=0.5\linewidth]{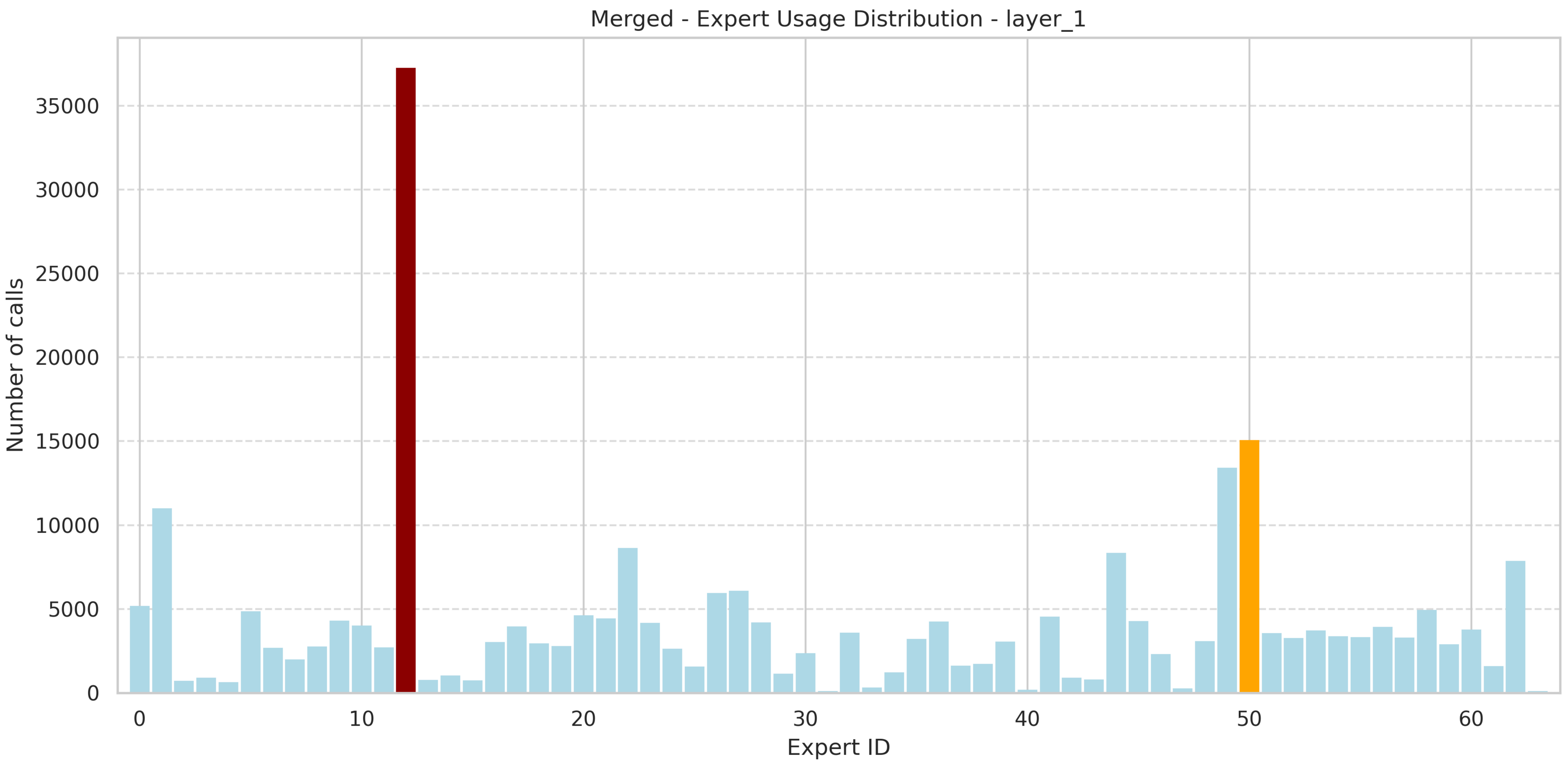}
    \vspace{-8pt}
    \caption{The activation frequency for the first layer. }
    \label{layer1}
    \vspace{-8pt}
\end{figure}

\begin{table*} [!ht]
\centering
\caption{Custom Variables Definition in Our Work}
\vspace{-8pt}
\begin{tabular}{c|c}
\toprule[1pt]
\textbf{Custom variables} & \textbf{Interpretation} \\
\midrule[1pt]
\textbf{Imbalance Score} & Evaluation of whether the number of times experts are called is balanced \\
\textbf{Expert Utilization} & The proportion of experts used \\
\textbf{Entropy (Normalized)} & Metrics to measure the uncertainty of model decisions \\
\textbf{Gini Coefficient} & Measuring the fairness or inequality of a distribution \\
\bottomrule[1pt]
\end{tabular}
\vspace{-8pt}
\label{variable_define}
\end{table*}

\begin{figure}[!h]
    \centering
    \includegraphics[width=0.98\linewidth, height=0.5\linewidth]{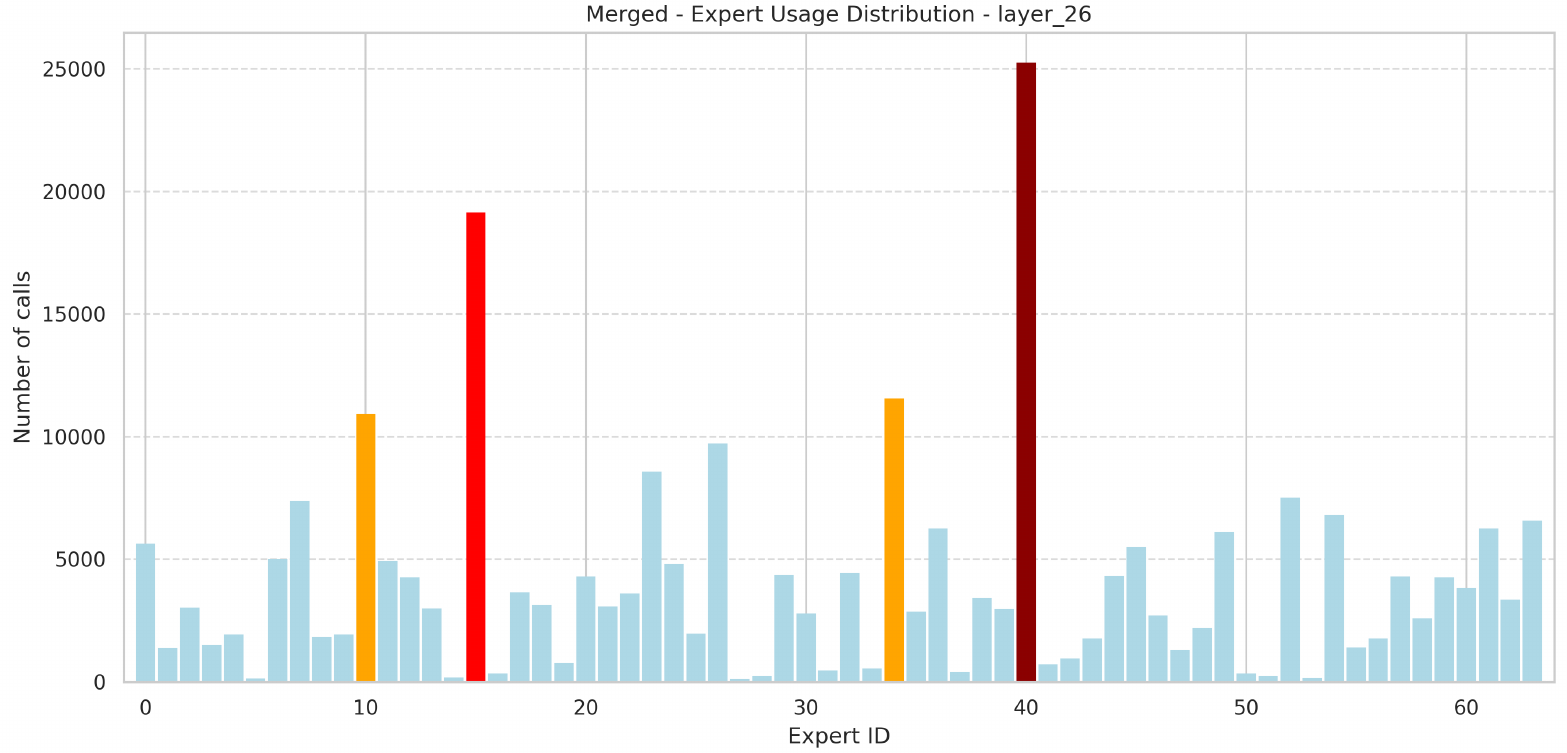} 
    \vspace{-8pt}
    \caption{The activation frequency for layer-26. }
    \label{layer26}
    \vspace{-8pt}
\end{figure}

\textbf{Activated Frequency for the Experts.} After the model processes each input problem, we record the six experts selected in each layer along with their corresponding weights. This approach enables detailed tracking of expert utilization across layers and facilitates analysis of how the model allocates tasks under different reasoning scenarios. We aggregate the selection frequency of each expert across all questions and analyze the usage patterns for each layer, as illustrated in Fig. \ref{layer1} and Fig. \ref{layer26}. This analysis provides insights into which experts play a more significant role during the reasoning process and serves as foundational data for subsequent error sensitivity experiments. Specifically, we present the usage frequencies of the experts in the first and 26th layers after processing all questions in Fig. \ref{layer1} and Fig. \ref{layer26}, respectively.

 
\begin{figure}[!h]
    \centering
    \includegraphics[width=0.98\linewidth, height=0.6\linewidth]{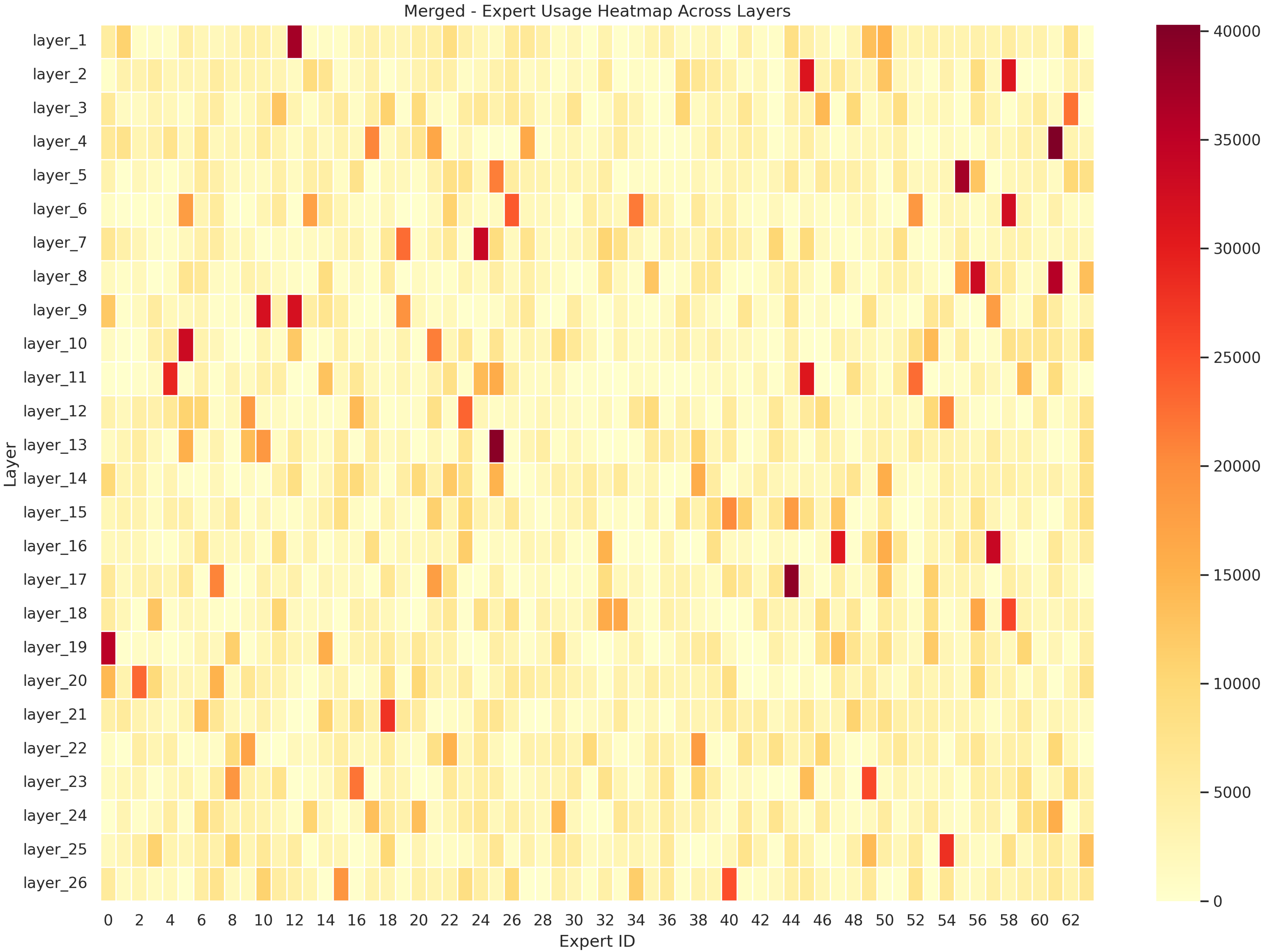}
    \vspace{-8pt}
    \caption{Heat map of activation frequency of experts. }
    \label{heatmap}
    \vspace{-8pt}
\end{figure}

\textbf{Activation Frequency Heatmap for the Experts. } To provide a more intuitive visualization, we present in Fig. \ref{heatmap} a heat map depicting the activation frequency of all experts across all layers following the input of the questions. The heat map reveals a pronounced imbalance in expert utilization: approximately 10 experts are activated more than 35,000 times throughout the entire reasoning process, whereas the majority of other experts are selected far less frequently, with some being used fewer than 10,000 times. This observation indicates that, despite the model comprising a large number of experts, only a small subset plays a central role in most reasoning tasks (e.g., expert 12 in layer 1). These results suggest that, for mathematical reasoning problems, a limited number of experts are responsible for the majority of computational tasks, leading to a marked "concentration" phenomenon in expert utilization.



This uneven distribution of expert activation provides important insights for error sensitivity analysis for the experts. We can reasonably hypothesize that introducing errors into the frequently activated experts will have a more pronounced impact on the model’s reasoning outcomes, as these experts predominantly drive the reasoning process. Experts that are activated less frequently, while still contributing to the overall model architecture, exert a relatively minor influence on overall performance when subjected to errors due to their limited participation. However, it is important to note that certain low-frequency experts (e.g., expert-0 in layer 1) also play critical roles in handling specific mathematical reasoning steps within particular tasks. Randomly setting the parameter values of expert-0 can lead to failures in managing these specialized scenarios, thereby disrupting the entire reasoning process. Although expert-0 is utilized far less than other experts, its involvement in key reasoning tasks renders it indispensable; errors in such experts can result in incorrect model outputs for those specific cases. Therefore, balancing the roles and resource allocation among experts—avoiding over-reliance on a small subset of high-frequency experts while ensuring that the potential of low-frequency experts is fully leveraged—emerges as a crucial consideration in our subsequent model optimization efforts. Finally, we define and track several customized variables to quantify and illustrate changes in expert activation across different layers of the original model on the GSM8K dataset, with detailed explanations provided in Tab. \ref{variable_define}.

\begin{figure}[!h]
    \centering
    \includegraphics[width=0.7\linewidth, height=1.0\linewidth, angle=90]{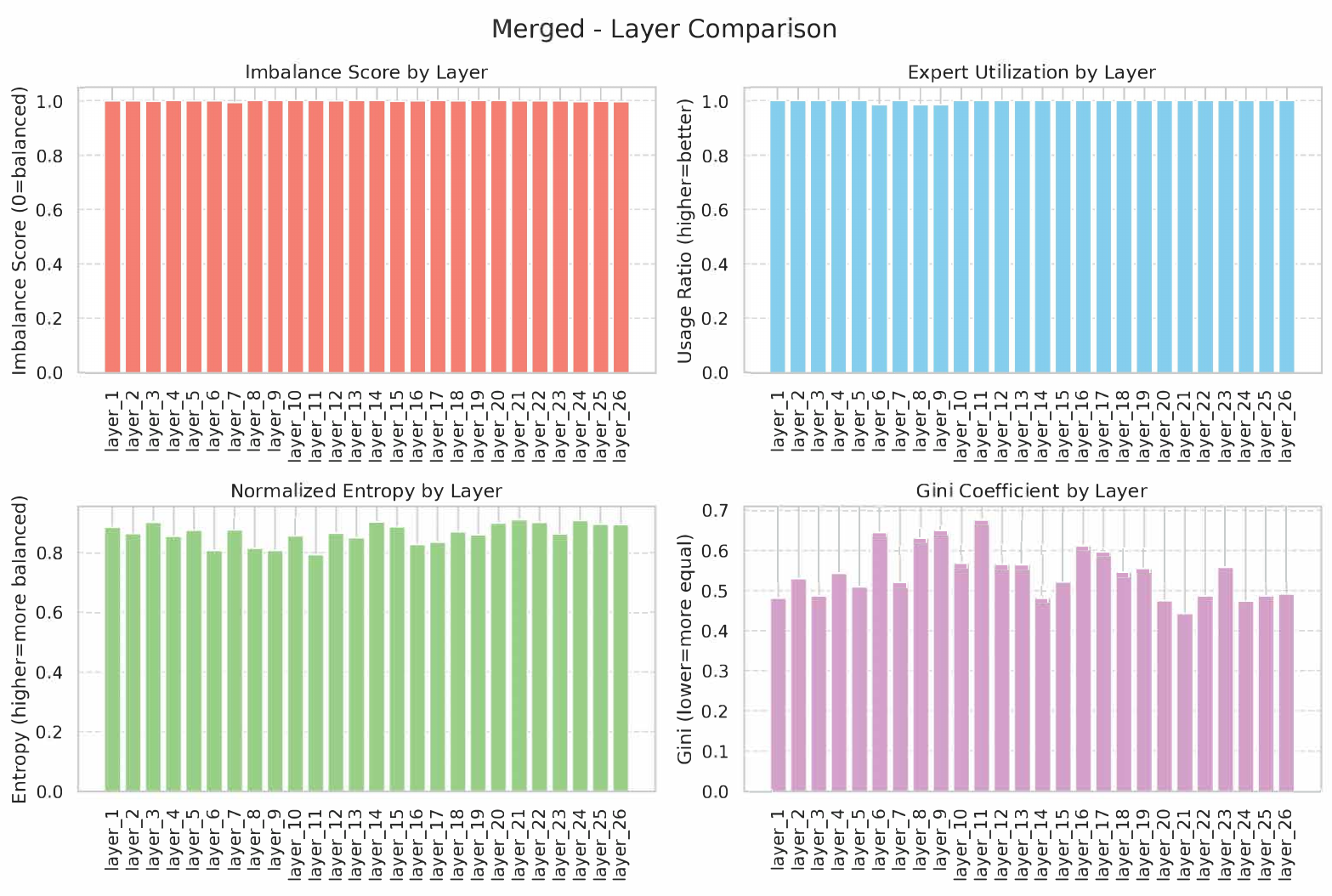}
    \vspace{-8pt}
    \caption{Comparison of aggregated expert layers }
    \label{expert_aggregate}
    \vspace{-8pt}
\end{figure}

As illustrated in Fig. \ref{expert_aggregate}, expert utilization across the 26 layers is highly imbalanced, with certain layers containing experts that remain inactive (as indicated by blue bar graph values less than 1.0). This observation is further corroborated by the normalized entropy and Gini coefficient metrics, both of which quantitatively confirm the uneven distribution of expert activation.


\subsection{Error Sensitivity Analysis for a Single Expert in Different Layers}

To investigate the impact of parameter errors on model inference accuracy, we conducted a series of experiments to evaluate the error sensitivity of different experts and provide a comprehensive analysis of the experimental results.



\subsubsection{Error Sensitivity Analysis for the Expert in the First Layer}

The distribution of the compression error for the currently most popular error-bounded lossy compression approaches \cite{liang2022sz3, huang2023cuszp, huang2024cuszp2} follows the Normal distribution $N\sim(0,\hat{e})$, where the $\hat{e}$ is the error bound of the compression approaches. Therefore, we randomly simulate $n$ errors that follow a Normal distribution with different error bounds $\hat{e}$ and add them to the experts in our experiments. 

Firstly, we choose the first expert in the first layer (expert-0) and add errors generated by different error bounds to it, including the error bound 
$\hat{e}=(10\%*\frac{||\theta_{\ell_{1}, expert_{0}}||_{1}}{n_{\ell_{1}, expert_{0}}})$, $\hat{e}=(30\%*\frac{||\theta_{\ell_{1}, expert_{0}}||_{1}}{n_{\ell_{1}, expert_{0}}})$, $\hat{e}=(50\%*\frac{||\theta_{\ell_{1}, expert_{0}}||_{1}}{n_{\ell_{1}, expert_{0}}})$, $\hat{e}=(80\%*\frac{||\theta_{\ell_{1}, expert_{0}}||_{1}}{n_{\ell_{1}, expert_{0}}})$, where the $\frac{||\theta_{\ell_{1}, expert_{0}}||_{1}}{n_{\ell_{1}, expert_{0}}}$ indicates the average value of the L1 norm of the expert-0 in the first layer, $n_{\ell_{1}, expert_{0}}$ is the number of parameters of the expert-0 in the first layer. The results indicate that introducing errors to expert 0 in the first layer does not affect the model's inference accuracy or the sequence of inference steps. This suggests that, for certain experts with minimal influence on overall model performance, adding small errors in their weights does not substantially impact the final output or the reasoning process. These findings demonstrate the model’s robustness to minor weight fluctuations, particularly when such changes do not involve critical experts. Key observations and analyses from these experimental results are summarized as follows:


$\bullet$ This phenomenon highlights a characteristic of the model: certain experts exert minimal influence on task outputs and may even be redundant. Consequently, modifying the weights of these experts does not significantly alter inference results. The model thus demonstrates high stability and resilience to small-scale errors or noise affecting these less critical experts. 
    
$\bullet$ However, when we randomized the weights of the first expert in the first layer, the results differed markedly. Although the model was still able to complete the inference task and generate an output, the inference result was entirely incorrect. This finding indicates that, even when the parameters of a single expert are severely corrupted, the model’s structure remains operational; however, the accuracy of its reasoning is significantly compromised. These results underscore that the contribution of seemingly unimportant experts in the reasoning process should not be underestimated. In certain tasks, where such experts may play a more substantial role than initially anticipated, parameter errors can directly lead to a dramatic decline in overall model performance.


$\bullet$ Although the model exhibits a degree of fault tolerance to parameter errors in individual, less influential experts, the role of each expert remains critical. Errors in expert parameters can substantially impact reasoning outcomes and diminish overall accuracy. Therefore, when designing MoE models, it is essential to ensure the correctness and stability of each expert to preserve the integrity and effectiveness of the overall reasoning process.


\subsubsection{Error Sensitivity Analysis for Highest-Frequently-Activated Expert}

We selected the most frequently activated expert in the first layer—specifically, expert 12, whose activation rate was substantially higher than that of other experts within the same layer. Same as the above section, we simulate the error for the expert with different error bounds $\hat{e}$ and add them to the expert, including the error bound ($\hat{e}$)
$\hat{e}=(30\%*\frac{||\theta_{\ell_{1}, expert_{26}}||_{1}}{n_{\ell_{1}, expert_{26}}})$, $\hat{e}=(50\%*\frac{||\theta_{\ell_{1}, expert_{26}}||_{1}}{n_{\ell_{1}, expert_{26}}})$, $\hat{e}=(80\%*\frac{||\theta_{\ell_{1}, expert_{26}}||_{1}}{n_{\ell_{1}, expert_{26}}})$, where the $\frac{||\theta_{\ell_{1}, expert_{26}}||_{1}}{n_{\ell_{1}, expert_{26}}}$ indicates the average value of the L1 norm of the highest-frequently-activated expert (the expert-26 in our benchmark) in the first layer, $n_{\ell_{1}, expert_{26}}$ is the number of parameters of the highest-frequently-activated expert (the expert-26 in our benchmark) in the first layer.




It is important to note that instruction compliance directly influences the overall effectiveness of the system. Even if the model's internal reasoning is correct, violations of output format or content constraints—referred to as non-instructional errors—can still lead to a reduction in system accuracy. Therefore, it is necessary to distinguish between two evaluation metrics as follows:


\begin{itemize}
    \item \textbf{Instruction Compliance Accuracy (ICA): }the output results meet both content correctness and format specifications. We abbreviate it as ICA in our work.
    \item \textbf{Pure Inference Accuracy (PIA): } only evaluate content correctness (ignoring format requirements). We abbreviate it as PIA in our work.
\end{itemize}

Experimental results indicate that as the error amplitude increases, the accuracy of reasoning with instruction compliance (i.e., the accuracy of the system output) gradually declines, although it remains relatively high overall. Notably, the model’s pure reasoning accuracy remains largely unaffected when errors are introduced exclusively to the most frequently activated expert in layer 1 (expert 26). The detailed results are presented in Tab. \ref{highest_frequent_expert}. A summary of the findings and corresponding analysis is provided below:


\begin{table}[!h]
\centering
\caption{The inference comparison for involving different errors in highest-frequently-activated expert (expert-26) in layer-1. ICA: Instruction Compliance Accuracy; PIA: Pure Inference Accuracy}
\vspace{-8pt}
\begin{tabular}{c|ccccc}
\toprule[1pt]
\textbf{Error Bound} & & \textbf{ICA} & &\textbf{PIA} &\\
\midrule[1pt]
Baseline & &0.86 & &0.96 &\\
$\hat{e}=(30\%*\frac{||\theta_{\ell_{1}, expert_{26}}||_{1}}{n_{\ell_{1}, expert_{26}}})$  & &0.82 & &0.96 &\\
$\hat{e}=(50\%*\frac{||\theta_{\ell_{1}, expert_{26}}||_{1}}{n_{\ell_{1}, expert_{26}}})$ & &0.80 & &0.96 &\\
$\hat{e}=(80\%*\frac{||\theta_{\ell_{1}, expert_{26}}||_{1}}{n_{\ell_{1}, expert_{26}}})$ & &0.79 & &0.95 &\\
\bottomrule[1pt]
\end{tabular}
\label{highest_frequent_expert}
\vspace{-8pt}
\end{table}

\begin{enumerate}
    \item \textbf{Adaptive protection of routing mechanism:} When the parameters of high-frequently-activated experts are distorted, the model dynamically adjusts its routing weights to reallocate tasks to other experts with intact functionality. This adaptive mechanism enables the model to preserve its core reasoning capabilities when involving errors.
    
    \item \textbf{Decoupling characteristics of instruction compliance and reasoning capabilities:} Parameter errors primarily disrupt output conventions—such as the box\{\} format required by instructions—rather than the underlying semantic generation capabilities, confirming the heterogeneity between the parameter space and the functional space in the MoE model.
    
\end{enumerate}

\subsubsection{Error Sensitivity Analysis for the Highest Frequently Activated Expert in Different Layers}


To assess the generalizability of the aforementioned conclusions—specifically, whether the model’s robustness to single expert parameter noise (error) is consistent across different layers—we conducted a series of controlled experiments targeting key layers throughout the model’s depth. In the 13th, 20th, and final (26th, output decision) layers, we selected the experts with the highest activation rates within each layer as the subjects for parameter perturbation. Same as the above sections, we set error bound as $\hat{e}=(80\%*\frac{||\theta_{\ell_{x}, expert_{y}}||_{1}}{n_{\ell_{x}, expert_{y}}})$, where $\ell_{x}$ indicate the $x-th$ layer and $expert_{y}$ indicate the expert$-y$. We examined the trends in both pure reasoning accuracy and instruction compliance accuracy of the model under cross-layer noise (error), and compared these results with those obtained from the first layer (Layer 1). The detailed experimental outcomes are presented in Tab. \ref{highest-expert-diff-layer}.



\begin{table}[!h]
\centering
\caption{The inference accuracy comparison for involving error in the highest-frequently-activated expert in different layers. ICA: Instruction Compliance Accuracy; PIA: Pure Inference Accuracy}
\vspace{-8pt}
\begin{tabular}{c|ccccc}
\toprule[1pt]
\textbf{Expert} & &\textbf{ICA} & & \textbf{PIA} &\\
\midrule[1pt]
\textbf{Baseline} && 0.86 && 0.96 &\\
\textbf{Layer1 (Expert-26)} && 0.79 && 0.95 &\\
\textbf{Layer13 (Expert-25)} && 0.75 && 0.94 &\\
\textbf{Layer20 (Expert-2)} && 0.89 && 0.96 &\\
\textbf{Layer26 (Expert-40)} && 0.96 && 0.96 &\\
\bottomrule[1pt]
\end{tabular}
\label{highest-expert-diff-layer}
\vspace{-8pt}
\end{table}

The experimental results demonstrate that the model’s core reasoning ability for mathematical problems remains highly stable (with accuracy consistently $ \geq 94\%$, even when errors are introduced to the most frequently activated experts. This suggests that the model’s semantic generation mechanism possesses strong resilience to the noise. In contrast, the ability to follow instructions exhibits a pronounced non-hierarchical progression: introducing errors into shallow layers (e.g., layer 1 and layer 3) significantly impairs this ability (by $10\% - 20\%$), whereas introducing errors into deeper layers (e.g., layer 20 and layer 26) leads to performance gains (of $7\% - 10\%$), likely due to implicit model integration effects. These findings reveal a functional decoupling between different expert layers (or experts) within the MoE architecture with respect to mathematical (semantic generation) logic and instruction parsing. Specifically, semantic reasoning is distributed and encoded throughout the entire network, while instruction compliance depends on specific layers. Moreover, adding noise to parameters in deeper layers can enhance task execution. This insight offers a novel perspective for the hierarchical design of robust MoE models.


\subsection{Error Sensitivity Analysis for Top$-k$ Most Frequently Activated Experts in a Layer}


Following the parameter error sensitivity analysis of the most frequently activated expert within a single layer, we extended the scope of our experiments to systematically investigate the impact of the noise (error) on the Top$-K$ most frequently activated experts in a given layer. Specifically, this experiment targets the 6 most frequently activated experts in both the first layer (Layer 1) and the decision layer (Layer 26). Same as the above section, we generated the error with the error bound of $\hat{e}=(80\%*\frac{||\theta_{\ell_{x}, expert_{y}}||_{1}}{n_{\ell_{x}, expert_{y}}})$, where $x=1,26$, $y$ indicates the top-6 frequently activated experts in layer 1 and layer 26, to simulate the change in reasoning ability when multiple experts are disturbed at the same time in actual scenarios. The experiment results are shown in Tab. \ref{topk-single-layer}.



\begin{table}[!h]
\centering
\caption{The inference accuracy for involving error in Top-K frequently activated experts in different layers. ICA: Instruction Compliance Accuracy; PIA: Pure Inference Accuracy}
\vspace{-8pt}
\begin{tabular}{c|ccccc}
\toprule[1pt]
\textbf{Layers} && \textbf{ICA} && \textbf{PIA} &\\
\midrule[1pt]
\textbf{Baseline} && 0.85 && 0.96 &\\
\textbf{Layer 1 Top-6 Experts} && 0.74 && 0.90 &\\
\textbf{Layer 26 Top-6 Experts} && 0.90 && 0.93 &\\
\bottomrule[1pt]
\end{tabular}
\label{topk-single-layer}
\end{table}

The results of experiments indicate that introducing errors into the top-$k$ most frequently activated experts leads to more pronounced performance degradation compared to adding errors to a single expert. Specifically, the instruction compliance rate in the first layer decreased from $79\%$  to $74\%$, while in the final layer it dropped from $96\%$ to $90\%$, as shown in Tab. \ref{topk-single-layer}. These findings suggest a cumulative effect when multiple key experts are simultaneously affected by errors. Nevertheless, it is noteworthy that the instruction compliance accuracy, even after introducing errors to the top 6 experts in the final layer, remains higher than that of the original model. This suggests that adding noise to the expert in the decision-making layer can enhance instruction compliance through an implicit integration effect. This mechanism facilitates the automatic generation of a diverse ensemble during the reasoning process, thereby improving output robustness. These results further support the notion that experts in the deep layers possess greater resilience to interference or may exhibit an inherent error compensation mechanism.


In distributed systems, our experimental findings offer a foundation for the quantitative analysis of fault-tolerant design in MoE modules. Specifically, it is essential to protect shallow, high-frequently-activated experts, whereas experts in deep layers can leverage their inherent redundancy to preserve functional integrity. Additionally, with respect to actual reasoning accuracy, we observed declines in both shallow and deep layers, with the impact being more pronounced in the first layer than in the final layer. This phenomenon provides preliminary evidence that the propagation of expert errors exhibits a cascade-like nonlinear amplification effect.


\subsection{Error Sensitivity Analysis for All Experts in a Single-Layer}


To further investigate the robustness of the model under large-scale parameter noise (error), we introduced errors to all 64 experts within individual layers. Same as the above sections, we generated errors with error bound $\hat{e}=(80\%*\frac{||\theta_{\ell_{x}, expert_{y}}||_{1}}{n_{\ell_{x}, expert_{y}}})$, where $x=1,13,20,26$, $y$ indicate all experts in layer 1, layer 13, layer 20 and layer 26. The experimental results are shown in Tab. \ref{all_experts_single_layer}.


\begin{table}[!h]
\centering
\caption{The Inference Accuracy for Involving Error for All Experts in a Single Layer. ICA: Instruction Compliance Accuracy; PIA: Pure Inference Accuracy}
\vspace{-6pt}
\begin{tabular}{cc|ccccc}
\toprule[1pt]
&\textbf{Layers}& & \textbf{ICA} && \textbf{PIA} &\\
\midrule[1pt]
&\textbf{Baseline} && 0.86 && 0.96 &\\
&\textbf{Layer1} && 0.33 && 0.71 &\\
&\textbf{Layer13} && 0.38 && 0.65 &\\
&\textbf{Layer20} && 0.61 && 0.80 &\\
&\textbf{Layer26} && 0.85 && 0.90 &\\
\bottomrule[1pt]
\end{tabular}
\label{all_experts_single_layer}
\end{table}

The experimental results presented in Tab. \ref{all_experts_single_layer} demonstrate that when all experts within a single layer are involved in errors, the instruction compliance ability of all models declines sharply, with compliance in shallow layers dropping to approximately $30\%$. Concurrently, we observed a decrease in mathematical reasoning performance across all models. However, the decline was less pronounced in deeper layers. This can be attributed to the proximity of deep layers to the output layer, which facilitates answer integration and output formatting according to directives. Additionally, the nonlinear amplification effect of expert data errors further contributes to these observations. For layer 13, the same experimental procedure resulted in the lowest inference accuracy among all layers. This finding suggests that, within the 26-layer network architecture, the middle layers play a critical role in problem analysis and reasoning.



\subsection{Error Sensitivity Analysis for High-Frequently-Activated Experts in a Group of Layers}


To further investigate the impact of errors on final inference accuracy, we partitioned the 26 expert layers of the Moonlight model into groups of 10 layers each, designating layers 17 to 26 as the final group. Within each group, we identified the most frequently activated expert and introduced errors to these experts, resulting in adding errors to a total of 10 experts. To simulate the impact of different errors on the model, we set the expert model parameter error bound as ($\hat{e}$)
$\hat{e}=(30\%*\frac{||\theta_{\ell_{x}, expert_{y}}||_{1}}{n_{\ell_{x}, expert_{y}}})$, $\hat{e}=(50\%*\frac{||\theta_{\ell_{x}, expert_{y}}||_{1}}{n_{\ell_{x}, expert_{y}}})$, $\hat{e}=(80\%*\frac{||\theta_{\ell_{x}, expert_{y}}||_{1}}{n_{\ell_{x}, expert_{y}}})$, where $\ell_{x}$ indicate the $x-th$ layer and $expert_{y}$ indicate the expert-y. The results are shown in Tab. \ref{group_error}.


\begin{table}[!h]
\centering
\caption{The result of involving errors in the highest-frequently-activated experts in a group of layers. Note: "/" indicates that the model no longer outputs or enters a down state. ICA: Instruction Compliance Accuracy; PIA: Pure Inference Accuracy.}
\vspace{-6pt}
\begin{tabular}{c|cc|cc|cc}
\toprule[1pt]
\multirow{2}{*}{ Grouping} & \multicolumn{2}{c|}{30\%} & \multicolumn{2}{c|}{50\%} & \multicolumn{2}{c}{80\%} \\ \cline{2-7}
 & ICA & PIA & ICA & PIA & ICA & PIA \\
 \midrule[1pt]
Group1 (L1-L10) & 0.86 & 0.94 & 0.81 & 0.91 & / & / \\
Group2 (L9-L18) & 0.81 & 0.89 & 0.69 & 0.75 & / & / \\
Group3 (L17-L126) & 0.91 & 0.95 & 0.92 & 0.94 & / & / \\
\bottomrule[1pt]
\end{tabular}
\label{group_error}
\end{table}

The results of the aforementioned experiments indicate that introducing an error with error bound of $\hat{e}=(80\%*\frac{||\theta_{\ell_{x}, expert_{y}}||_{1}}{n_{\ell_{x}, expert_{y}}})$ caused all model groups to fail to generate valid outputs. This finding suggests that adding large errors to the most frequently activated experts across multiple layers exerts a substantially greater impact than a similar operation in single-layer architectures, potentially resulting in complete model failure. In instances where valid outputs were still produced, further analysis revealed that errors most severely affected the intermediate layers, corroborating our earlier conclusion that these layers are primarily responsible for the model’s core reasoning processes. For shallow layers, although errors were propagated and amplified through nonlinear cascading effects, their main influence was on the attention mechanism, leading the model to disproportionately focus on irrelevant tokens and thereby degrading overall performance. In contrast, the deeper layers exhibited strong robustness—even when multiple expert components were perturbed—and, in some cases, demonstrated enhanced instruction-following capabilities, highlighting their role in output generation and representation. In summary, the propagation and impact of parameter errors are primarily determined by the functional role of the affected layer, rather than following a simple monotonic relationship with network depth.


\subsection{Generalization Evaluation of Error Sensitivity for the Experts}

We further conducted experiments to analyze the error sensitivity of the model using the mathematics dataset \cite{hendrycks2021measuring}. The primary objective of this experiment is to evaluate the performance and stability of the MoE model across different datasets, as well as to investigate its robustness when errors are introduced by various experts.

\begin{figure}[!h]
    \centering
    \includegraphics[width=0.98\linewidth, height=0.68\linewidth]{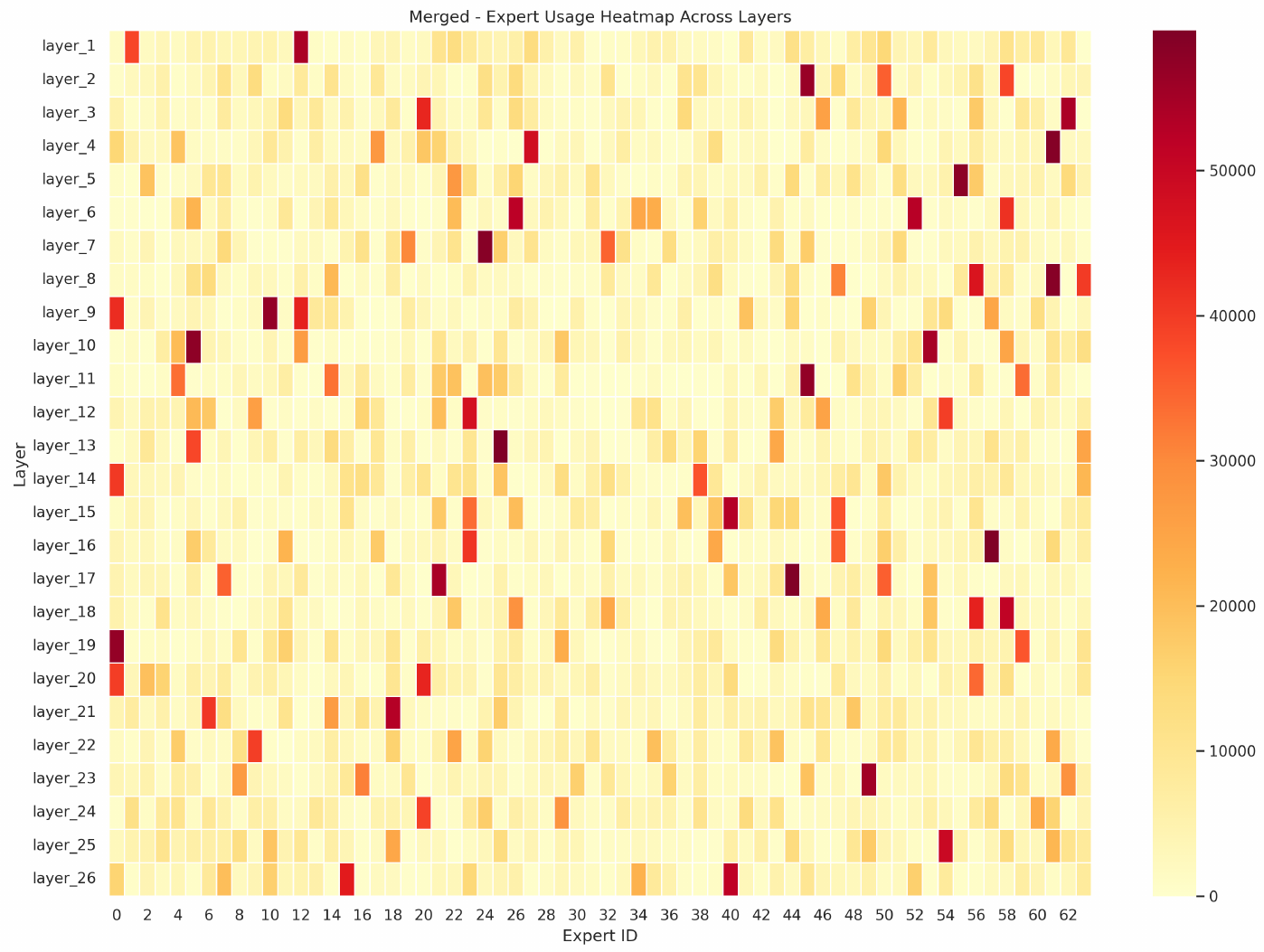}
    \vspace{-8pt}
    \caption{Heat map of activation frequency of experts for the Math dataset.}
    \label{Distribution}
\end{figure}

First, we present a heatmap illustrating the activation frequency of different experts on the mathematics dataset in Fig. \ref{Distribution}. Our observations indicate that, for more challenging mathematical datasets, the number of highly activated experts increases substantially. This suggests that complex tasks necessitate the engagement of a greater number of experts to address heightened computational demands and data complexity. However, despite this increase, the distribution of high-frequency expert activation remains inhomogeneous. Specifically, while certain experts play a significantly enhanced role in particular domains or task characteristics, the overall distribution of expert activation is still uneven across other areas. This inhomogeneity implies that, although expanding the number of experts can improve the model’s capacity to process complex data, it does not necessarily lead to balanced optimization across all tasks.


\vspace{-8pt}

\subsubsection{Error Sensitivity Analysis for the Highest-Frequently-Activated Expert in Different Layers based on the Math Dataset}


To validate the generalizability of the preliminary conclusions drawn in the preceding sections, we selected the most frequently activated experts in layers 1, 13, 20, and 26 for further experimentation in this section. We also set the error bound as $\hat{e}=(80\%*\frac{||\theta_{\ell_{x}, expert_{y}}||_{1}}{n_{\ell_{x}, expert_{y}}})$, where $\ell_{x}$ indicate the $x-th$ layer and $expert_{y}$ indicate the expert$-y$. The experiment results are shown in the following Tab. \ref{math_dataset}.


\begin{table}[!h]
\centering
\caption{Results of involving errors in the highest frequently activated experts in different layers based on the Math dataset. ICA: Instruction Compliance Accuracy; PIA: Pure Inference Accuracy.}
\vspace{-6pt}
\begin{tabular}{cc|ccccc}
\toprule[1pt]
&\textbf{Layers} && \textbf{ICA} && \textbf{PIA} &\\
\midrule[1pt]
&\textbf{Baseline} && 0.62 && 0.70 &\\
&\textbf{Layer1} && 0.60 && 0.65 &\\
&\textbf{Layer13} && 0.56 && 0.60 &\\
&\textbf{Layer20} && 0.62 && 0.68 &\\
&\textbf{Layer26} && 0.66 && 0.70 &\\
\bottomrule[1pt]
\end{tabular}
\label{math_dataset}
\vspace{-8pt}
\end{table}

The results indicate that, overall, the introduction of errors into experts leads to a decline in most mathematical and logical reasoning abilities. Notably, when errors were injected into the expert at layer 13, the model’s instruction-following accuracy and pure reasoning accuracy dropped significantly to 0.56 and 0.60, respectively. This substantial performance degradation suggests that middle-layer experts are particularly sensitive and play a critical role in the model’s language understanding and decision-making processes. In contrast, the instruction-following accuracy for the expert at layer 26 increased slightly to 0.66, implying that experts in deep layers contribute to output generation in a more redundant or robust manner. Additionally, the impact of errors on experts in layers 1 and 20 was relatively minor, with the performance of the 20th layer remaining nearly identical to the baseline.


Overall, the experiment further confirms the uneven distribution of expert activations and the varying importance of different layers within the model. Even when errors of identical magnitude are introduced, the impact of highly activated experts on model performance is heterogeneous across layers. This finding indicates that, although the number of frequently activated experts increases, their actual contribution is determined by the functional role of their respective layers. Consequently, future research should focus on joint modeling approaches that incorporate hierarchical semantics and expert sensitivity in expert scheduling and model architecture optimization, as these strategies are likely to enhance the robustness and generalization capabilities of large-scale models.


\subsubsection{Error Sensitivity Analysis for High-Frequently-Activated Experts in a Group of Layers based on the Math Dataset}

In this section, we will further explore the impact when involving the errors in high-freqently-activated experts in a group of layers based on the math dataset. Same as the above sections, we set the error bound as $\hat{e}=(30\%*\frac{||\theta_{\ell_{x}, expert_{y}}||_{1}}{n_{\ell_{x}, expert_{y}}})$ and $\hat{e}=(50\%*\frac{||\theta_{\ell_{x}, expert_{y}}||_{1}}{n_{\ell_{x}, expert_{y}}})$, where $\ell_{x}$ indicate the $x-th$ layer and $expert_{y}$ indicate the expert$-y$. The experiment results are shown in the following Tab. \ref{math_dataset2}.


\begin{table}[!h]
\centering
\caption{The inference accuracy for involving errors in most highly activated experts in a group of layers. ICA: Instruction Compliance Accuracy; PIA: Pure Inference Accuracy}
\vspace{-8pt}
\begin{tabular}{c|cc|cc}
\toprule[1pt]
\multirow{2}{*}{Grouping} & \multicolumn{2}{c|}{30\%} & \multicolumn{2}{c}{50\%} \\ \cline{2-5}
 & ICA & PIA & ICA & PIA \\
 \midrule[1pt]
Group1 (L1-L10) & 0.60 & 0.63 & 0.50 & 0.58 \\
Group2 (L9-L18) & 0.58 & 0.62 & 0.43 & 0.50 \\
Group3 (L17-L26) & 0.63 & 0.65 & 0.56 & 0.60 \\
\bottomrule[1pt]
\end{tabular}
\label{math_dataset2}
\end{table}
\vspace{-6pt}

Based on the above results, it is evident that for more challenging mathematical problems, involving errors for experts leads to a more pronounced decline in model performance, affecting both instruction-following accuracy and pure reasoning ability. Furthermore, the findings indicate that the middle layers are primarily responsible for the model’s core logical reasoning tasks, whereas the experts in deep layers facilitate final integration and output generation. In contrast, experts in the shallow layers mainly handle the transformation of input tokens into vector representations. When the parameters of these shallow experts are compromised, the model tends to focus on less relevant tokens, thereby amplifying errors throughout the subsequent reasoning process.


\section{Takeaways}

In this study, we investigate the impact of compression errors in different experts on final inference accuracy from seven distinct perspectives: \ding{172} introducing errors to a single expert within a specific layer; \ding{173} introducing errors to the most frequently activated expert in a given layer; \ding{174} introducing errors to the most frequently activated experts across different layers; \ding{175} introducing errors to the top-$k$ most frequently activated experts; \ding{176} introducing errors to all experts within various layers; \ding{177} introducing errors to the most frequently activated expert within a group of layers; and \ding{178} evaluating the aforementioned scenarios across different benchmark datasets. Based on the experimental results, we summarize nine key conclusions as follows:

\ding{182} The introduction of parameter errors to a single expert does not substantially affect the model’s reasoning performance, either in terms of instruction-following ability or actual reasoning capability. However, when the parameters of an expert are fully randomized, the model experiences severe degradation, with a marked decline in reasoning ability. This outcome suggests that even experts perceived as less important play a critical role in maintaining overall model functionality.


\ding{183} For frequently activated experts, even if their errors are large (e.g., $\hat{e}=(50\%*\frac{||\theta_{\ell_{x}, expert_{y}}||_{1}}{n_{\ell_{x}, expert_{y}}})$ and $\hat{e}=(80\%*\frac{||\theta_{\ell_{x}, expert_{y}}||_{1}}{n_{\ell_{x}, expert_{y}}})$, where $\ell_{x}$ indicate the $x-th$ layer and $expert_{y}$ indicate the expert-y), the model is able to maintain high reasoning accuracy, demonstrating a degree of robustness in the presence of mild errors. However, these results also indicate that the introduction of errors to experts primarily impacts the model’s instruction-following ability before affecting its reasoning performance.


\ding{184} When errors are introduced to high-frequency experts in each layer in increasing order, both instruction-following and reasoning performance exhibit a non-monotonic distribution pattern. This indicates that the functional contributions of experts at different layers within the MoE architecture—specifically in semantic generation and instruction parsing—are not sequentially distributed. Instead, these contributions manifest at various hierarchical levels and collectively influence the model’s overall performance.


\ding{185} When errors are introduced into the high-frequently-activated experts across different layers, we observe that the middle-layer experts are primarily responsible for the core reasoning tasks of the model. In contrast, experts in the shallow layers predominantly manage the attention mechanism and the transformation of input tokens into vector representations, while those in the deep layers are mainly involved in instruction following and output integration. In summary, the impact of parameter errors is determined by the primary function of the affected layer, rather than by a simple progression with increasing depth.


\ding{186} Through systematic experimentation, we elucidate the mechanism by which expert parameter errors within the MoE influence the inference performance of large-scale models and derive several key application principles. Specifically, in the context of industrial-scale model maintenance and security protection, particular attention should be directed toward the shallow and middle layers. The shallow layers are responsible for the initial deployment of the attention mechanism, while the middle layers perform core reasoning functions. The integrity of parameters in these layers is therefore critical, as it directly impacts the accuracy of model reasoning.


\ding{187} In specific applications, we observed that, when processing mathematical datasets, adding errors in the shallow layers has minimal impact on model performance; in some instances, the performance following shallow layer interference even surpasses that of the middle layer. This can be attributed to the relatively low demand for detailed contextual understanding in mathematical problems, where the output from shallow experts is generally sufficient to provide the necessary information. However, in more complex tasks—particularly those requiring a nuanced understanding of meaning, context, and implicit information—shallow experts play a crucial role in the initial processing of information.


\ding{188} For experts in the deep layers, which are primarily responsible for instruction following and result generation, the introduction of controllable data errors during training and inference can serve as an effective optimization strategy. This approach has the potential to enhance the overall robustness and accuracy of the model. Notably, in tasks that demand high-dimensional feature abstraction, the incorporation of appropriate levels of error can facilitate improved generalization, thereby enabling the model to perform more effectively on complex tasks.


\ding{189} Experts in the shallow layers, which are primarily responsible for the attention mechanism and the transformation of input tokens into vector representations, exhibit minimal degradation in inference accuracy when subjected to bounded errors. In contrast, errors in the middle-layer experts, which are central to model reasoning, significantly impair inference accuracy. Interestingly, introducing bounded errors in the deep-layer experts, which are mainly responsible for instruction following and output integration, can sometimes lead to improvements in inference accuracy.



\section{Conclusion}
Mixture-of-Experts (MoE) models have enabled large language models (LLMs) to transition from dense architectures to those featuring sparsely activated experts. However, deploying MoE models within the constraints of limited GPU memory presents significant challenges. Offloading strategies have emerged as effective solutions for reducing GPU memory overhead, but they introduce additional costs associated with transferring experts between GPU memory and main memory. Consequently, efficient compression of experts has become a critical issue for the practical implementation of offloading strategies. In this work, we provide a comprehensive analysis of how compression errors affect inference accuracy, considering scenarios involving errors in both single and multiple experts. We conduct extensive experiments across various benchmarks to systematically evaluate the impact of introducing errors into different experts. Our results indicate that experts in the shallow layers are primarily responsible for the attention mechanism and the transformation of input tokens into vector representations, while experts in the deep layers are mainly tasked with instruction following and output integration.


\section{Acknowledgment}
We thank the reviewers for their insightful suggestions. This work was supported in part by the National Natural Science Foundation of China under Grant 62302420, in part by the U.S. Department of Energy, Office of Science, Advanced Scientific Computing Research (ASCR) under Grant DE-AC02-06CH11357, and in part by the Hong Kong Innovation and Technology Support Programme (ITP) under Grant ITP/012/25LP.


\clearpage
\bibliographystyle{ACM-Reference-Format}
\bibliography{acmsc25_workshop}

\appendix

\end{document}